\documentclass[journal,twoside,web]{IEEEtran}
\usepackage{cite}
\usepackage{amsmath,amssymb,amsfonts}
\usepackage{mathtools}
\usepackage{graphicx}
\usepackage{hyperref}
\usepackage{textcomp}
\usepackage{graphicx}
\usepackage{tabularx}
\usepackage[font=small,labelfont=bf]{caption}
\usepackage[numbib]{tocbibind}
\usepackage{url}
\usepackage[section]{placeins}
\usepackage[shortlabels]{enumitem}
\usepackage{algpseudocode}
\usepackage{svg}
\usepackage{tabularray}
\usepackage{float}
\usepackage{tikz}
\usepackage{orcidlink}
\usepackage{subcaption}
\usetikzlibrary{positioning}
\usepackage{pgfplots}
\pgfplotsset{compat=1.16}
\usepackage{xcolor}
\usepackage{multirow}
\usepackage[dvipsnames]{xcolor}

\def\BibTeX{{\rm B\kern-.05em{\sc i\kern-.025em b}\kern-.08em
    T\kern-.1667em\lower.7ex\hbox{E}\kern-.125emX}}

\newcommand\Tstrut{\rule{0pt}{3ex}}
\newcommand\Bstrut{\rule[-1.5ex]{0pt}{0pt}}

\newcommand{\customarrowbullet}{%
  \tikz[baseline=-0.5ex]{
    \draw[->] (-0.1em,0.4em) -- (0,0) -- (1em,0);
  }}

\title{SweetDeep: A Wearable AI Solution for Real-Time Non-Invasive Diabetes Screening}
\author{Ian Henriques \orcidlink{0009-0007-3756-9734}, Lynda Elhassar \orcidlink{0009-0004-9502-6840}, Sarvesh Relekar \orcidlink{0009-0005-7377-7843}, Denis Walrave, Shayan Hassantabar \orcidlink{0000-0003-4297-2097}, Vishu Ghanakota, Adel Laoui, Mahmoud Aich, Rafia Tir, Mohamed Zerguine, Samir Louafi, Moncef Kimouche, Emmanuel Cosson \orcidlink{0000-0002-8785-3385}, Niraj K Jha \orcidlink{0000-0002-1539-0369}, \IEEEmembership{Fellow, IEEE}
\thanks{Ian Henriques and Niraj K. Jha are with the Department of Electrical and Computer Engineering, Princeton University, Princeton, NJ 08544, USA (e-mail: ih2422@princeton.edu; jha@princeton.edu).}
\thanks{Lynda Elhassar, Sarvesh Relekar, Denis Walrave, Vishu Ghanakota, and Adel Laoui are with the AI company NeuTigers, Inc. (e-mail: lynda@neutigers.com; sarvesh@neutigers.com; denis@neutigers.com; shayan@neutigers.com; vishu@neutigers.com; adel@neutigers.com).}
\thanks{Mahmoud Aich and Rafia Tir are self-practicing endocrinologists in Paris, France (email: m.aich@free.fr; raphia.t@hotmail.fr).}
\thanks{Mohamed Zerguine and Emmanuel Cosson are with AP-HP Avicennes, Bobigny, France (email: mohamed.zerguine@aphp.fr; emmanuel.cosson@aphp.fr). Samir Louafi and Mouncef Kimouche are with Clinic International Constantine, Constantine, Algeria (email: s.louafi@issgroupfrance.com; moncef.kimouche@ciconstantine.com).}
\thanks{\textbf{This work has been submitted to the IEEE for possible publication. Copyright may be transferred without notice, after which this version may no longer be accessible.}}}
\begin{document}
\maketitle

\noindent
\begin{abstract}
The global rise in type 2 diabetes underscores the need for scalable and cost-effective screening methods. Current diagnosis requires biochemical assays, which are invasive and costly. Advances in consumer wearables have enabled early explorations of machine learning-based disease detection, but prior studies were limited to controlled settings. We present SweetDeep, a compact neural network trained on physiological and demographic data from 285 (diabetic and non-diabetic) participants in the EU and MENA regions, collected using Samsung Galaxy Watch 7 devices in free-living conditions over six days. Each participant contributed multiple 2-minute sensor recordings per day, totaling approximately 20 recordings per individual. Despite comprising fewer than 3,000 parameters, SweetDeep achieves 82.5\% patient-level accuracy (82.1\% macro-F1, 79.7\% sensitivity, 84.6\% specificity) under three-fold cross-validation, with an expected calibration error of 5.5\%. Allowing the model to abstain on less than 10\% of low-confidence patient predictions yields an accuracy of 84.5\% on the remaining patients. These findings demonstrate that combining engineered features with lightweight architectures can support accurate, rapid, and generalizable detection of type 2 diabetes in real-world wearable settings.
\end{abstract}


\section{Introduction}

\IEEEPARstart{D}{iabetes} is a common metabolic disorder characterized by impaired glucose regulation, increasing the risk of heart disease, stroke, and kidney failure \cite{singh2025type}. As of 2025, diabetes affects over 589 million people worldwide \cite{idf2025} (including around 38 million in the U.S. \cite{cdc_diabetes_statistics}), many of whom are unaware they have the condition. Type 2 diabetes -- the most common type, accounting for around 90\% of cases -- is influenced by genetics, age, and lifestyle factors, such as physical activity and nutrition \cite{singh2025type}. Given its growing prevalence and impact \cite{idf2025}, improving early detection of type 2 diabetes in at-risk populations is a critical public health priority.

Detecting type 2 diabetes usually involves some combination of three biochemical tests \cite{singh2025type}: fasting plasma glucose (FPG), oral glucose tolerance testing (OGTT), and glycated hemoglobin (A1c). These require blood samples, can be costly and time-consuming, and sometimes yield inconsistent results \cite{tucker2020limited}, making diagnosis particularly challenging and expensive. Since the three tests rely on specialized laboratory procedures, diabetes is tested less frequently than necessary, contributing to substantial underdiagnosis (as high as 42.8\% of all global cases) \cite{idf2025}. As a result, many early-stage type 2 cases that could still be treated or reversed remain undetected.

While biochemical approaches remain the clinical gold standard for diagnosing type 2 diabetes in the U.S. \cite{american20242} and internationally \cite{idf2025}, the disease also exhibits well-established physiological correlates -- such as blood pressure \cite{teuscher1989diabetes} and heart rate variability \cite{benichou2018heart} -- that can be measured non-invasively and at low cost. Multiple prior studies \cite{yin2019diabdeep, cordeiro2021hyperglycemia, susana2022non} have leveraged these physiological markers to train machine learning models that estimate diabetes risk and approximate glucose levels in the laboratory. Each referenced study used one of three non-invasive sensor modalities: bioelectrical analysis (BIA), electrocardiography (ECG), or photoplethysmography (PPG). These sensors are available in recent commercial smartwatches \cite{vashist2019commercially}, highlighting the potential for wearable AI solutions capable of detecting type 2 diabetes in patients’ homes.

However, existing physiological models for detecting type 2 diabetes have shown limited validation in real-world settings. Most are trained and evaluated under highly controlled conditions (e.g., fixed timing of data collection, constrained movement, regulated mealtimes, and exclusion of comorbidities or concomitant medications), and their performance degrades in everyday, naturalistic contexts \cite{zanelli2022diabetes, piet2023non, moses2023non}. As a result, developing robust, non-invasive detection models that generalize across laboratory environments, diverse populations, and real-world conditions remains an ongoing research problem.

We address this challenge by collecting a large dataset for machine learning that combines features carefully extracted from BIA, ECG, and PPG signals with family history and age questionnaire responses, across a diverse group of 285 participants from the EU and MENA regions. Of these, 162 are ND (Non-Diabetic) and 123 are T2D (Type 2 Diabetic).

In place of laboratory measurements, we adopt a regulatory decentralized clinical trial (DCT) approach \cite{van2021decentralized}, where participants record data at home using smartwatches (Samsung Galaxy Watch 7). Each participant is instructed to take 2-minute sensor recordings (\textit{instances}) six times per day, before and after meals (without strict meal timing) to capture diurnal signal variations. All individuals are included in the study, regardless of disease comorbidities or prescribed medications.

We apply this new dataset to train \textit{SweetDeep} -- a small, edge-optimized neural network that leverages non-invasive physiological signals, demographic information, and temporal context to predict each participant’s diabetic status (ND vs. T2D) in real time. SweetDeep is trained to separate the two classes, and subsequently evaluated using three-fold cross-validation testing, with an inter-patient split to prevent data leakage between folds. We report instance-level and patient-level accuracies, along with macro-F1, sensitivity, specificity, and calibration error. We also introduce a post-training \textit{abstention} (``Don't Know" class \cite{herbei2006classification}) adapter that refuses to classify under 10\% of study participants with low-confidence predictions at test time. Such cases can be referred for standard clinical testing, ensuring safety while maintaining higher accuracy on confident predictions. SweetDeep makes progress towards improving cost-effective type 2 diabetes screening in primary care, telehealth, and consumer health settings.

Our main contributions are summarized as follows.
\begin{itemize}
    \item We prepare a new, large-scale and diverse dataset with 400+ participants across multiple countries, ages, medication groups, and diabetic status (ND, T2D, and Pre-Diabetic) to enable real-world model generalization.
    \item We apply signal processing techniques and smartwatch application programming interfaces (APIs) to extract clinically relevant features from noisy smartwatch signals (ECG, PPG, and BIA), enabling representations that are independent of exact sensor design.
    \item Our model, SweetDeep, achieves 82.5\% patient-level accuracy for classification of T2D versus ND, with balanced predictions (a macro-F1 of 82.1\%), high specificity (84.6\%), and good sensitivity (79.7\%).
    \item Test-time results from SweetDeep suggest that the model is well-calibrated, with a low expected calibration error of 5.5\% at the patient level (5.4\% at the instance level), offering reliable probabilistic estimates of diabetes status.
    \item Our abstention modification utilizes calibration to reject low-confidence patient predictions at test time, resulting in an accuracy of 84.5\% on the remaining patients.
\end{itemize}

The remainder of the paper is organized as follows. In Section \ref{related}, we discuss related work. We describe in detail our DCT protocol, data collection process, dataset construction, model training, and additional implementation details in Section \ref{methods}. Section \ref{metrics} describes performance evaluation metrics used in our study, while Section \ref{results} presents our model results, both at the instance and patient levels. Section \ref{conclusion} presents concluding remarks and future directions for our research.

\section{Related Work} \label{related}

This section surveys prior work in biochemical diagnostic assays, wearable sensing for diabetes detection, and neural network reliability. 
We first describe biochemical "gold standard” tests used to assign patient labels in our study, followed by prior approaches for estimating diabetes risk using non-invasive biophysical sensors. The section concludes with a discussion of strategies to improve the reliability of probabilistic predictions, with an emphasis on model calibration.

\subsection{The Biochemical ``Gold Standard"}

According to clinical ``gold standard" criteria provided by the American Diabetes Association (ADA) \cite{american20242} and the International Diabetes Federation (IDF) \cite{bergman2024international}, type 2 diabetes is diagnosed by combining three invasive biochemical assays.

The first of these, FPG, involves drawing a blood sample to measure an individual's glucose levels in \textit{fasting} conditions \cite{american20242}. 
An FPG reading exceeding 126 mg/dL indicates diabetes, while FPG measurements below 100 mg/dL are considered normal (non-diabetic). FPG values between 100 mg/dL and 126 mg/dL indicate that an individual may be \textit{pre-diabetic}, trending towards type 2 diabetes but not yet diabetic.

Next, the OGTT uses chemical tests performed on blood samples to track \textit{post-prandial} (post-meal) fluctuations in glucose levels caused by ingestion of a sugar solution \cite{bergman2024international}. OGTT has two variants, 1-hour and 2-hour, both of which have specific thresholds (in mg/dL) for classifying individuals as non-diabetic, pre-diabetic, or diabetic \cite{american20242, bergman2024international}.

Finally, the A1c assay provides a longer-term measure of glycemic control by estimating average blood glucose levels over the preceding 8-12 weeks, based on the extent of glucose binding (glycation) to hemoglobin in red blood cells \cite{nathan2007relationship}. A1c values below 5.7\% are considered non-diabetic, while values above 6.5\% indicate diabetes \cite{american20242}.

According to ADA and IDF guidelines, a patient is classified as diabetic if \textit{any} of the three assays yields a diabetic-range result, and as non-diabetic if \textit{all} assays fall within the non-diabetic range \cite{american20242, bergman2024international}. We adopt this convention to assign “ground truth” labels of diabetic status (ND or T2D) to patients in our dataset. Our smartwatch-based machine learning framework is designed to infer these clinically established labels directly from non-invasive physiological signals, enabling rapid type 2 diabetes screening without costly laboratory testing.

\subsection{Biophysical Sensing Alternatives}

Unlike traditional biochemical tests, wearable \textit{biophysical sensors} provide a non-invasive means of monitoring physiological activity (e.g., heart rate and blood pressure) using electrical or optical probes. 
Although typically less specific and more susceptible to interference compared to highly-controlled biochemical assays, these sensors offer significant potential for predicting and identifying physiological ``fingerprints" associated with diseases like diabetes.

In particular, type 2 diabetes is frequently accompanied by Cardiac Autonomic Neuropathy (CAN) \cite{pop2010cardiac, duque2021cardiovascular}, a cardiovascular condition resulting from damage to the autonomic nerves that regulate the heart and blood vessels. CAN progressively develops -- even in patients receiving anti-diabetic treatment -- as chronic hyperglycemia leads to glycation of glucose to proteins and lipids \cite{pop2010cardiac, azad1999effects}. Early manifestations of CAN can often be detected within the first year of diabetes onset by examining heart rate variability (HRV) and the corrected QT interval (QTc), both of which can be derived from biophysical ECG signals \cite{pop2010cardiac}. Consequently, T2D individuals often exhibit an ECG-based physiological “fingerprint,” proportional to disease severity, which is distinct from that of ND individuals.

A related study used ECG signals to detect elevated blood glucose levels \cite{cordeiro2021hyperglycemia}. In that work, the authors extracted several fiducial features -- including HRV metrics and QTc -- to train a machine learning model that classified laboratory-acquired ECG segments as hyperglycemic (above 100 mg/dL) or normoglycemic \cite{cordeiro2021hyperglycemia}. While the referenced study provided valuable evidence that ECG features can reflect glucose dysregulation, it adopted a non-standard threshold for hyperglycemia (typically defined as $>$180 mg/dL \cite{singh2025type}) and used an intra-patient split, in which samples from the same individuals appeared in both the training and test sets. Such an evaluation design can overestimate performance by ``leaking" patient-specific physiological patterns to the test set \cite{chaibub2019detecting, tougui2021impact}. Our study uses similar ECG-derived features to reflect the progression of CAN in diabetic patients, but we also use clinically validated diagnostic labels, smartwatch-acquired ECG signals collected outside controlled laboratory settings, and an \textit{inter-patient split} to promote generalization across individuals.

In addition to ECG, PPG sensing provides a complementary biophysical signal for detecting T2D. PPG uses a colored light source and associated photodetector to estimate blood volume changes during heartbeats, enabling assessment of cardiovascular health, including HRV and arterial stiffness parameters \cite{susana2022non}. Similar to ECG, prior research has demonstrated that PPG-derived features can distinguish diabetic from non-diabetic individuals in laboratory settings, e.g., classifying ``diabetes” (conservatively defined as $>$200 mg/dL from random glucose readings) against healthy controls in real-time \cite{susana2022non}.

In practice, PPG sensors are highly prone to motion artifacts, especially in free-living conditions \cite{castaneda2018review}. To mitigate this, we align PPG samples with stable ECG segments for motion-free measurements. Using built-in smartwatch APIs, we extract 10 PPG-derived features related to arterial blood pressure, which complement CAN-related ECG indicators by reflecting diabetes-linked vascular activity \cite{brands1996poor, wu2022association}.

Recently, DiabDeep \cite{yin2019diabdeep} applied a machine learning algorithm trained on PPG and other smartwatch signals -- including Galvanic Skin Response and skin temperature -- together with questionnaire data, to detect type 2 diabetes in real-time. While DiabDeep demonstrated that smartwatches can capture physiological indicators of diabetes, its dataset included only 13 T2D and 27 ND individuals in tightly controlled laboratory conditions. In addition, DiabDeep's evaluation process involved an intra-patient split, further limiting generalization of its results to larger, more diverse patient groups.

Our work extends DiabDeep and the other mentioned studies by combining biophysical sensors available on Samsung Galaxy Watch 7 -- including ECG, PPG, and BIA -- with a risk factor questionnaire, enabling accurate neural network prediction of diabetes status in a decentralized fashion across larger and more diverse populations.

\subsection{Model Reliability}

To ensure that models like SweetDeep produce clinically reliable predictions, prior studies have emphasized the importance of examining probabilistic behavior and calibration.

Most neural network classifier models report multiple probabilities on the output softmax layer, reflecting the model's \textit{confidence} that an instance belongs to each class. Given a set of class-wise predicted probabilities $\hat p(y=i | x)$ for a single instance, a two-class model would typically select the ``correct class" as the one with the highest probability:
\begin{equation}
    \hat y(x) \coloneqq \arg\max_{i \in \mathcal \{0, 1\}} \hat{p}(y=i|x)
\end{equation}

This formulation suffices in cases where we are only concerned with accuracy. However, in clinical applications, it is helpful to use predicted probabilities to determine how much the model's ``correct" output should be trusted. For instance, 
$\hat{p}(y=1|x) = 55\%$ 
would suggest that the model's output (class 1) is too unconfident to be taken seriously. Meanwhile, a different $\hat{p}(y=1|x) = 5\%$ would suggest that the true label for that instance is almost certainly class 0.

For predicted probabilities to be clinically meaningful, they must be \textit{well-calibrated}, i.e., reported probabilities should accurately reflect the true likelihood of correctness. For example, a well-calibrated model would ensure that among all instances with $\hat{p}(y=1|x) = 0.7$, approximately 70\% are positive cases.

Neural networks trained with standard cross-entropy loss are often overconfident and poorly calibrated, making their probabilistic estimates unreliable \cite{guo2017calibration}. Post-hoc calibration techniques can improve the reliability of probabilistic predictions, but they require validation data (reducing the effective training set) and may fail when training and validation distributions differ, as is common with physiological signals \cite{guo2017calibration, ovadia2019can}.

To address the calibration problem, we leverage a known trend in the literature: larger networks (with greater width and depth) tend to exhibit higher calibration errors, developing emergent ``overconfidence" on new test data \cite{guo2017calibration}. Accordingly, we adopt a compact architecture to limit overconfidence, with shallow depth (just two hidden layers) and limited width (at most 64 neurons per layer). In Section~\ref{size-var}, we show that further increasing SweetDeep's model size may degrade calibration without improving accuracy.

\section{Methodology} \label{methods}

This section describes the methodology underlying dataset and model construction. We first present the architecture and key components of the SweetDeep real-time inference framework. Subsequently, we detail procedures for data acquisition and physiological sensor quality control. Next, we outline the feature extraction approach, which integrates hand-engineered features with smartwatch API-derived metrics. Finally, we describe the SweetDeep model, including the training recipe and rules for patient-level prediction and abstention.

\begin{figure*}[!ht]
    \centering
    \includegraphics[width=\textwidth]{inference_colored.pdf}
    \caption{\sf An overview of the SweetDeep real-time inference framework, consisting of three main stages: signal and questionnaire \textcolor{Cerulean!70!black}{data collection} with quality control, \textcolor{Orchid}{neural network inference} (with two hidden layers), and optional \textcolor{Goldenrod!60!black}{confidence-based abstention}.}
    \label{fig:inference}
\end{figure*}

\subsection{Framework Overview}

Fig.~\ref{fig:inference} shows the proposed framework for SweetDeep real-time inference. For each \textit{data instance} (sensor recording session), 
smartwatch signals are converted into a small set of engineered features, which is merged with patient-specific questionnaire data (family history and age) to create a 35-dimensional feature vector. This vector is fed to the neural network, which produces two outputs: the T2D class probability $\hat p(y = 1|x)$ and ND class probability $\hat p(y = 0|x)$. For each patient, these probabilities are averaged across multiple data instances, enabling more accurate \textit{patient-level} predictions that incorporate physiological data from different times of day.

Based on these probabilities, the model may decide (with fixed thresholds) that the prediction is too unconfident, in which case it abstains on classification by reporting the label ``Don't Know" while referring the user to more comprehensive biochemical testing. Otherwise, the diabetes screening result (consisting of the predicted class and probabilities) is immediately reported to the user and relevant medical professionals.

\subsection{Data Collection Protocol}

The training data for SweetDeep were collected in a decentralized manner from 438 study participants over one year. Each participant contributed six consecutive days of recordings ($\sim$20 recordings per individual). The data collection protocol was led by AP-HP Avicennes, the largest hospital network in France, with similar coordination in the MENA region via the CIC network in Algeria. In parallel, data collection has begun in the United States to support future validation studies.

The study resulted in the diagnosis of 162 ND, 153 Pre-Diabetic (PD), and 123 T2D patients,
highlighting the prevalence of diabetes in the general population. The majority of T2D patients were already taking diabetes-specific medication, but 24 new T2D cases were also discovered. All patients were included in our study, regardless of medication for comorbidities including hypertension and high cholesterol. PD patients were excluded from the model training and evaluation
to focus on ND/T2D comparisons, but their data are preserved for later experiments.
Clinical labels were determined for each patient by applying the aforementioned biochemical ``gold standard," combining FPG,
OGTT-1h, OGTT-2h, and A1c assays, as recommended by the ADA and IDF \cite{american20242, bergman2024international}.

Sensor traces and questionnaire responses were gathered from participants through two mobile applications: SweetDeep Collect (on the Samsung Galaxy Watch 7)
and SweetDeep Manage (on the Samsung Galaxy S20 phone), respectively. 

The SweetDeep Collect App used each watch's sensor logging module -- with access to Samsung's Privileged Health SDK -- to read raw
ECG, PPG, BIA, Blood Oxygen, Skin Temperature, and Heart Rate signals. In particular, ECG recordings were manually initiated by patients before and after each meal
(about six times per day), and other measurements (including time of day) were synchronized against ECG. Watch signals were locally stored for the duration
of the study, then transferred to remote servers in Europe for model training.

Meanwhile, the SweetDeep Manage eCRF (electronic Case Report Form) App was used by medical staff to enroll patients, store clinical labels, check the status
of data collection, stop data collection, and monitor overall progress and high-level statistics. It was also used to streamline application installations
on the watch. Risk-related questionnaires (pertaining to age, family history, and other factors) were presented once per patient, and saved for later model training.

\subsection{Sensor Quality Control}

Although study participants were instructed to start ECG recordings while fully stationary, we found that high-voltage ``spike" artifacts (due to finger slips or wrist movements) still frequently appeared in traces. These artifacts cannot be remedied by filtering alone, and we find that isolating and processing \textit{stable segments} using strict quality assessment considerably improves the quality of extracted features.

\begin{figure}[t]
    \centering
    \includegraphics[width=\columnwidth]{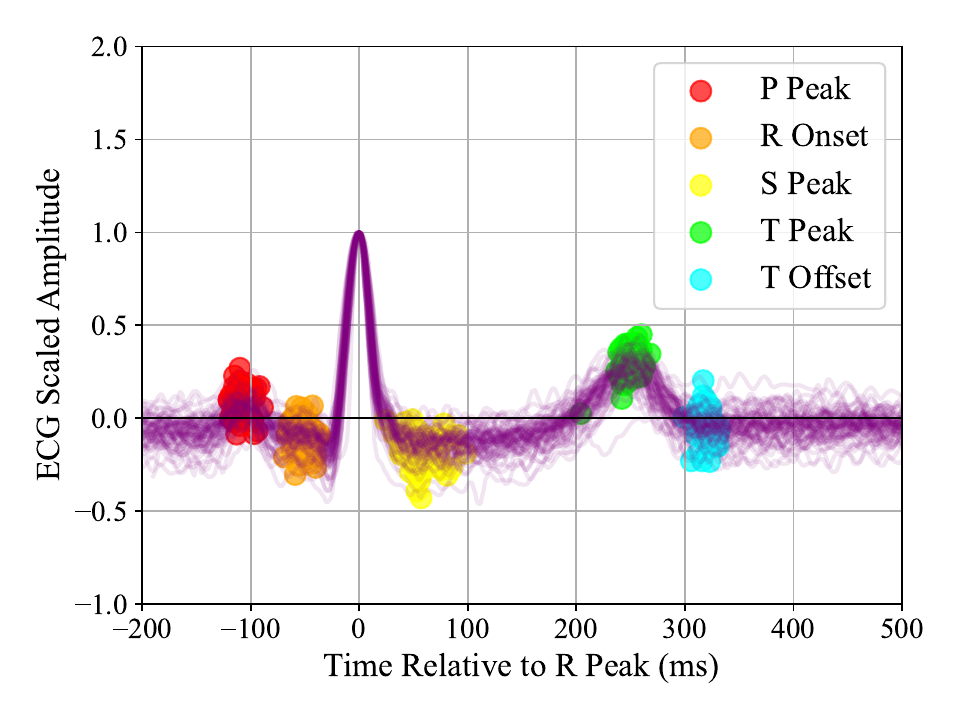}
    \caption{\sf Example delineation of many stable, filtered ECG beats from a single recording after quality control steps are applied. Notably, the \textcolor{orange}{R Onsets} and \textcolor{cyan}{T Offsets}, two components of QTc, have consistent positions across heartbeats.}
    \label{fig:ecg}
\end{figure}

Using a peak-finding approach, we align several stable (i.e., spike-free), filtered ECG heartbeats by their R peaks (maxima), then apply wavelet-based delineation to locate other critical points (including Q troughs and T peaks). We follow this step with outlier-based and clinical threshold-based filtering on relevant intervals (including QT and RR) to eliminate highly abnormal heartbeats. 
The resulting cleaned beats are stable and consistent enough (as can be seen from Fig.~\ref{fig:ecg}) to derive diabetes-related features including QTc and HRV.

If fewer than 10\% of heartbeats remain after quality control for an ECG signal, the corresponding data instance (i.e., all sensor modalities and questionnaire inputs associated with that recording) is excluded from the dataset. This prevents the model from learning spurious correlations between diabetes labels and signal artifacts arising from low-quality data. Instances with low-quality PPG or BIA signals and out-of-range features are similarly identified and discarded, to maintain high dataset integrity for model training and evaluation.

During SweetDeep real-time inference, we can apply similar quality checks to all sensors on the fly, to ensure that physiological signal readings meet the standards required for reliable neural network predictions. When data quality falls below some threshold (e.g., due to motion artifacts or sensor failure), the screening application would prompt users to retake ECG recordings or wait before attempting again.

\subsection{Clinical Feature Extraction}

For ECG, PPG, BIA recordings that pass quality control, it is essential to extract descriptive features that meaningfully support diabetes classification. Specifically, each sensor trace should be reduced to a compact set of temporal and morphological features that capture the underlying physiological “fingerprint” of diabetic versus healthy individuals.

While feature extraction is frequently automated using deep learning on raw signal traces, we instead employ clinically grounded features, including QTc, HRV, cardiac output (CO), and basal metabolic rate (BMR). These domain-specific features offer several advantages: They improve training efficiency by reducing task complexity, mitigate overfitting to spurious signal artifacts, enhance the interpretability of learned representations, and promote cross-device generalization to wearables with varying sensor configurations.

First, the ECG-derived \textit{corrected QT interval} (i.e., QTc) is known to increase in response to ventricular weakness, and it frequently rises above normal levels as a result of diabetic neuropathy conditions including CAN \cite{elming2002qtc, pop2010cardiac}. We calculate QTc using the Fridericia correction \cite{vandenberk2016qt}, to account for the influence of heart rate on the measured QT interval:
\begin{equation}
    \text{QTc} = \frac{\text{delay(R Onset, T Offset)}}{[\text{delay(R Peak, next R Peak)]}^{1/3}}
    \label{qtc}
\end{equation}

The delay between consecutive R Peaks in the denominator of Eq.~(\ref{qtc}), also known as the \textit{RR interval} (or equivalently, the inter-beat interval), appears again in the formulas for heart rate variability. It is well known that decreased variability between nearby beat interval lengths (HRV) can reflect autonomic nervous system damage, exposing another detectable signature of CAN \cite{electrophysiology1996heart}. We use two independent constructions of HRV \cite{electrophysiology1996heart} that are based on the variation of RR intervals:
\begin{align}
    &\text{SDNN} = \sqrt{\frac{1}{N-1} \sum_{i=1}^{N} (\text{RR}_i - \overline{\text{RR}})^2}\\
    &\text{RMSSD} = \sqrt{\frac{1}{M} \sum_{j=1}^{M} (\Delta\text{RR}_j)^2}
\end{align}
where $\overline{\text{RR}}$ is the mean RR interval across $N$ (not necessarily consecutive) samples in the recording. $\Delta{\text{RR}_j}$ is the difference between consecutive RR interval lengths, iterating over $M \le N-1$ such pairs. The combination of QTc, the Standard Deviation of RR Intervals (SDNN), and the Root Mean Square of Successive Differences (RMSSD) forms the \textbf{ECG group}, with three features.

From the PPG sensor, we extract 10 features using Samsung's Health Data SDK, corresponding to estimated Total Peripheral Resistance (TPR) and CO. These blood pressure-related features (the \textbf{PPG-BP group}) must be calibrated about once a month with a discrete blood pressure monitor, but remain accurate over the course of our study.

We use the same SDK to extract 10 metabolism-related features (the \textbf{BIA group}) from the bioelectrical impedance sensor, including BMR and Body Fat Mass (BFM). These features remain relatively stable throughout the day, serving as useful indicators of long-term metabolic health and lifestyle.

While extracted via smartwatch-specific APIs, the aforementioned PPG-BP and BIA features are based on standard sensors and signal processing methods, and can be replicated (with proper calibration and testing) on other wearable platforms equipped with PPG, BIA, and access to raw sensor data.

Questionnaire responses contribute four more relevant features. Specifically, continuous \textbf{Age} (in years) comprises one, and categorical \textbf{Family History} of diabetes is one-hot encoded into three features: 0 (no history), 1 (grandparent, uncle, aunt, or cousin), and 2 (parent, sibling, or child). The dataset thus has 27 features that cover cardiovascular, metabolic, demographic, and genetic dimensions of diabetes pathology.

\subsection{Time-of-Day Representation}

\begin{figure*}[t]
    \centering
    \includegraphics[width=\textwidth]{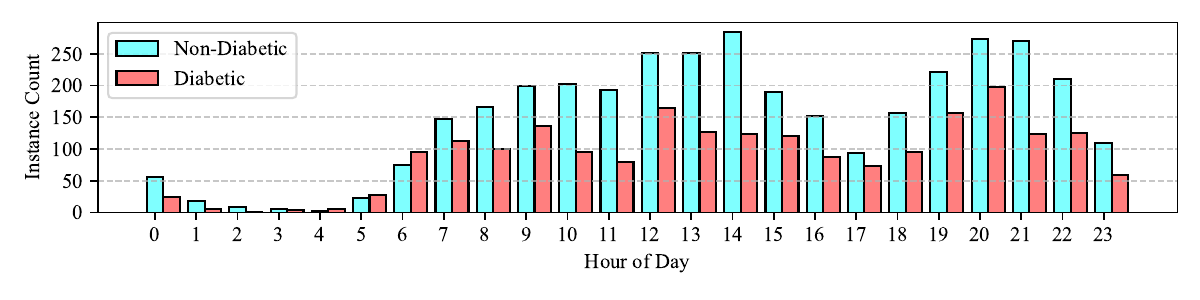}
    \caption{\sf Time-of-day counts (clustered by hour) for instances in \textcolor{cyan}{non-diabetic} and \textcolor{red}{diabetic} cohorts. Most hours have $\ge$50 instances from both cohorts, supporting the inclusion of time-of-day features to account for circadian variations in physiological signals.}
    \label{fig:time}
\end{figure*}

Many of the aforementioned feature categories exhibit well-documented circadian variation, characterized by predictable fluctuations over a 24-hour period. For example, arterial blood pressure typically rises in the early morning, peaks in the early afternoon, and declines toward night \cite{baumgart1991circadian}. In contrast, HRV metrics, such as SDNN, tend to reach their highest values in the early morning, decrease throughout the afternoon, and increase again during sleep \cite{huikuri1990reproducibility}.

As shown in Fig.~\ref{fig:time}, the dataset provides adequate coverage across most hours of the day for both type 2 diabetic and non-diabetic cohorts. We leverage this temporal diversity by augmenting the model input with a set of eight \textbf{Time} features, allowing us to directly account for the confounding influence of time on physiological measurements.

Specifically, time-of-day information is represented using \textit{sinusoidal encodings}, analogous to positional encodings in vanilla transformers \cite{vaswani2017attention} and temporal encodings in environmental and energy forecasting models \cite{eddin2023location, bansal2025temporal}. To our knowledge, this is the first application of sinusoidal temporal features in a tabular dataset for disease classification.

We first encode the time of each instance, in seconds relative to midnight ($t$), as a \textit{circadian phase} $\phi$ between $0$ and $2\pi$:
\begin{equation}
    \phi = \frac{2\pi t}{86400}
\end{equation}

From this circadian phase, we create eight Time features:
\begin{align*}
    \{&\sin(\phi), \cos(\phi), \sin(2\phi), \cos(2\phi),\\
       &\sin(3\phi), \cos(3\phi), \sin(4\phi), \cos(4\phi)\}
\end{align*}

This representation is cyclical: the phases of times 23:59 and 00:00 differ by nearly $2\pi$, producing almost identical feature values at the day boundary. Each successive pair of sinusoids oscillates at a higher frequency, enabling the model to capture increasingly fine-grained temporal context.

Overall, the combined feature set (35 features), including the time-of-day encodings, provides a compact input representation (see Table \ref{features}) that supports diabetes screening under free-living conditions with a smartwatch at any time of day.

\begin{table}[ht]
\centering
\normalsize
\renewcommand{\arraystretch}{1.2}
\scalebox{1.0}{
\begin{tabular}{|l|p{4.6cm}|}
\hline
\textbf{Feature Group} & \textbf{Features (Abbrev.)} \Tstrut\Bstrut \\
\hline
ECG & QTc, SDNN, RMSSD \Tstrut\Bstrut\\
\hline
PPG-BP & 8$\times$TPR, 2$\times$CO \Tstrut\Bstrut\\
\hline\Tstrut
BIA & {\raggedright BodMag, 4$\times$ConMag, BodDeg, BFM, SMM, TBW, BMR} \Tstrut\Bstrut \\
\hline
Age & Age (in years) \Tstrut\Bstrut \\
\hline
Family History & None, $\text{2}^\text{nd}$-degree, $\text{1}^\text{st}$-degree \Tstrut\Bstrut\\
\hline
Time & 8 sinusoidal encodings \Tstrut\Bstrut\\
\hline
\end{tabular}
}
\caption{\centering \sf The 35-feature set used to train SweetDeep.}
\label{features}
\end{table}

\subsection{Fold Creation and Model Training}

The SweetDeep dataset is split into three class-stratified folds (95 patients in each) using an \textit{inter-patient} split to prevent subject-level leakage. Features in each fold are min-max normalized on the training partition (excluding Time features to preserve their $[-1, 1]$ sinusoidal range), and SMOTE \cite{chawla2002smote} is applied to each training set to mitigate class imbalance.

SweetDeep is a fully connected network with 35 inputs, two hidden layers, with 64 and 8 units, and two softmax outputs. Each layer performs a linear transformation followed by 1D BatchNorm (BN) \cite{ioffe2015batch}, ReLU, and Dropout (DO; $p_\text{DO}=0.1$) \cite{srivastava2014dropout}. The model is trained using Adam \cite{kingma2014adam} with batch size 512 for 50 epochs (to reach $\sim$90\% training accuracy while limiting overfitting) without additional hyperparameter tuning.

\subsection{Prediction and Abstention}

At the data instance level, T2D predictions are given by the softmax probability $\hat p_x = \hat p(y=1|x)$, with a standard threshold of 0.5, enabling screening results after just one recording if desired. Meanwhile, patient-level predictions (also thresholded at $50\%$ probability) summarize the model's overall probabilistic judgment over all of a patient's measurements.

For a given participant $X$, let $\pi_X$ denote the unknown, underlying distribution of the data instances.  Assuming a sufficient number of instances that cover diverse times of day, each observed instance $x_i$ can be regarded as an independent and identically distributed sample from $\pi_X$. The true patient-level probability of a T2D outcome is then defined as the expected value of all instance-level probabilistic predictions:
\begin{equation}
    \hat p_X \coloneqq \mathbb{E}_{x \sim \pi_X} \big[\hat p_x\big]
\end{equation}

We approximate this expectation as the empirical mean over the $N$ observed instances for patient $X$:
\begin{equation}
    \hat p_X \approx \frac{1}{N} \sum_{i=1}^{N} \hat p_x
\end{equation}

In particular, when model probabilities $\hat p_x$ are well-calibrated at the instance level, patient-level predictions $\hat p_X$ are also well-calibrated, due to the linearity of expectation. We apply this calibration guarantee to implement an abstention protocol, where low-confidence predictions (i.e., $\hat p_X \approx 0.5$) are cautiously labeled as ``Don't Know.'' We apply the criterion
\begin{equation}
    |\hat p_X - 0.5| < 0.08
\end{equation}
to decide when to abstain. In practice, this rule abstains from classifying under 10\% of study participants across three folds.

\section{Performance Evaluation Metrics} \label{metrics}

This section provides a brief overview of the performance evaluation metrics used to assess SweetDeep, beginning with accuracy-related metrics (i.e., accuracy, specificity, sensitivity, and macro-F1) and ending with expected calibration error.

\subsection{Accuracy Measures}

SweetDeep is trained to assign a binary ``hard" label (ND or T2D) to each instance $x$ and patient $X$ by thresholding the predicted probability ($\hat p_x$ or $\hat p_X$) at 0.5. Predictions with $\hat p \ge 0.5$ are labeled T2D (\textit{positive}), while those with $\hat p < 0.5$ are labeled ND (\textit{negative}).

If a prediction matches the ground truth, it is \textit{True} -- either True Positive (TP) or True Negative (TN). Otherwise, it is \textit{False} -- False Positive (FP) or False Negative (FN). These four cases help define several standard performance metrics.

\textbf{Accuracy} is given by
\begin{equation}
    \text{Accuracy} = \frac{\text{TP} + \text{TN}}{\text{TP} + \text{FP} + \text{FN} + \text{TN}}
\end{equation}

Although widely used, accuracy can be misleading under class imbalance. For example, if 90\% of samples are ND, a trivial model that always predicts ND achieves 90\% accuracy yet fails to identify any diabetic patients, yielding a 100\% false negative rate with a 0\% `sensitivity' to the disease.

Formally, \textbf{sensitivity} quantifies detection of T2D cases:
\begin{equation}
    \text{Sensitivity} = \frac{\text{TP}}{\text{TP} + \text{FN}}
\end{equation}
while \textbf{specificity} measures how rarely false alarms occur:
\begin{equation}
    \text{Specificity} = \frac{\text{TN}}{\text{TN} + \text{FP}}
\end{equation}

An accuracy measure robust to class imbalance is \textbf{macro-F1} ($\mathcal F_1$) \cite{opitz2019macro}, which averages F1 scores across both classes:
\begin{equation}
\mathcal F_1 = 
\frac{1}{2} \left(
\frac{2\,\text{TP}}{2\text{TP} + \text{FP} + \text{FN}} +
\frac{2\,\text{TN}}{2\text{TN} + \text{FP} + \text{FN}}
\right)
\end{equation}

Together, these four complementary metrics provide a more reliable view of model performance than accuracy alone.

For model evaluation, a three-fold cross-validation strategy is employed to mitigate potential bias associated with a particular test set. To derive cross-validation performance measures, counts (TP, FP, FN, TN) are summed across all folds at the patient or instance level, to obtain aggregate binary classification results. Each reported accuracy, sensitivity, specificity, and $\mathcal F_1$ is calculated using the aggregated counts from the same three-fold data partitioning, enabling consistent comparison of metrics both within and across models.

\subsection{Calibration Error}

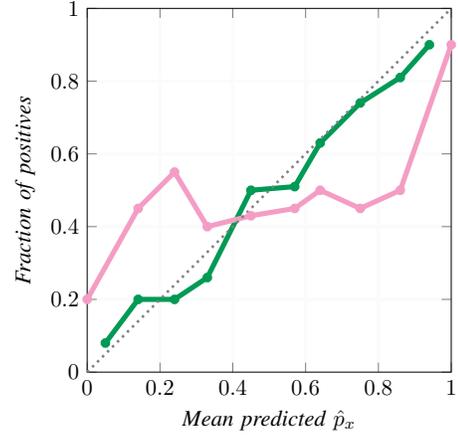
\begin{figure}[t]
    \centering
    \begin{tikzpicture}[scale=0.85]
        \begin{axis}[
            xlabel=\textit{Mean predicted $\hat{p}_x$},
            ylabel=\textit{Fraction of positives},
            grid=both,
            grid style={line width=0.5mm, draw=gray!3},
            xmin=0, xmax=1.0,
            ymin=0, ymax=1.0,
            xtick={0,0.2,...,1.0},
            ytick={0,0.2,...,1.0},
            axis equal image
        ]
        \addplot[mark=*,ForestGreen,line width=0.8mm,mark size=1,opacity=1] plot coordinates {
            (0.05, 0.08)
            (0.14, 0.20)
            (0.24, 0.20)
            (0.33, 0.26)
            (0.45, 0.50)
            (0.57, 0.51)
            (0.64, 0.63)
            (0.75, 0.74)
            (0.86, 0.81)
            (0.94, 0.9)
        };
        \addplot[mark=*,Lavender,line width=0.8mm,mark size=1,opacity=1] plot coordinates {
            (0.0, 0.2)
            (0.14, 0.45)
            (0.24, 0.55)
            (0.33, 0.4)
            (0.45, 0.43)
            (0.57, 0.45)
            (0.64, 0.5)
            (0.75, 0.45)
            (0.86, 0.5)
            (1.0, 0.9)
        };
        \addplot[dotted, gray, line width=0.4mm] plot coordinates {
            (0, 0)
            (1.0, 1.0)
        };
        \end{axis}
    \end{tikzpicture}
    \caption{\sf Toy calibration curves, with 10 evenly spaced bins. \\(\textcolor{ForestGreen}{green} = well-calibrated, \textcolor{Thistle}{pink} = poorly calibrated)}
    \label{fig:cal_toy}
\end{figure}

To assess the quality of ``soft" probabilistic predictions from SweetDeep, we use calibration error metrics to compare predicted T2D probabilities $\hat p$ with observed test outcomes.

Calibration is typically evaluated using a \textit{10-bin} approach, where predictions falling within the same interval -- e.g., $\hat p \in [0.7, 0.8)$ -- are grouped. The empirical frequency of diabetes within each bin ($\bar p_\text{bin}$) is then compared with the mean predicted probability ($\hat p_\text{bin}$) of T2D within that bin.

This relationship is often visualized with a calibration curve, where each bin-wise summary $(\hat p_\text{bin}, \bar p_\text{bin})$ forms a point, as exemplified by Fig.~\ref{fig:cal_toy}. A model whose calibration curve closely follows the dotted diagonal line $\bar p_\text{bin} = \hat p_\text{bin}$ is considered to be well-calibrated, indicating strong agreement between predicted and observed T2D probabilities in the testing set.

To make this observation more concrete, \textbf{Expected Calibration Error (ECE)} \cite{guo2017calibration} measures the average bin-wise residual between observed and predicted probabilities:
\begin{equation}
    \text{ECE} = \sum_{b \in \text{bins}} \frac{|b|}{|\mathcal D_\text{test}|} |\bar p_b - \hat p_b| 
\end{equation}
given $|b|$ samples in bin $b$ and $|\mathcal D_\text{test}|$ samples in the entire test set. Lower ECE indicates better calibration, with $10\%$ as a typical reference for well-calibrated models \cite{guo2017calibration}.

Similarly to accuracy, we use three-fold cross-validation to collectively bin test samples from all three folds, resulting in aggregate calibration curves and ECE values that reflect overall calibration performance of SweetDeep across the dataset.

\section{Experimental Data and Results} \label{results}

This section highlights our overall experimental findings after training and evaluating the SweetDeep dataset across all three folds. First, the accuracy and calibration performance of SweetDeep is studied. Then, ablations of the training approach, including architectural layer removals, larger model sizes, feature elimination, and threshold adjustments, are tested and compared against the unaltered SweetDeep model. The section concludes by analyzing cohort-specific trends and discussing results after an optional confidence-based abstention step.

\begin{table*}[ht]
    \normalsize
    \centering
    \makebox[\textwidth][c]{
        \begin{tabular}{!{\vrule width 1pt}c!{\vrule width 1pt}c!{\vrule width 1pt}c!{\vrule width 1pt}c!{\vrule width 1pt}c!{\vrule width 1pt}c!{\vrule width 1pt}}
        \noalign{\hrule height 1pt}
        \textbf{SweetDeep Result} & \textbf{Accuracy~(\%)} $\left[\uparrow\right]$ & \textbf{Sensitivity~(\%)} $\left[\uparrow\right]$ &
        \textbf{Specificity~(\%)} $\left[\uparrow\right]$ & $\boldsymbol{\mathcal{F}_1}$~\textbf{(\%)} $\left[\uparrow\right]$ &
        \textbf{ECE~(\%)} $\left[\downarrow\right]$ \Tstrut\Bstrut \\
        \noalign{\hrule height 1pt}
        \textcolor{Periwinkle}{Instance-Level} & 80.9 & 77.5 & 82.9 & 79.8 & \textbf{\textcolor{Periwinkle}{5.4}} \Tstrut\Bstrut \\
        \noalign{\hrule height 1pt}
        \textcolor{Fuchsia}{Patient-Level} & \textbf{\textcolor{Fuchsia}{82.5}} & \textbf{\textcolor{Fuchsia}{79.7}} & \textbf{\textcolor{Fuchsia}{84.6}} & \textbf{\textcolor{Fuchsia}{82.1}} & 5.5 \Tstrut\Bstrut \\
        \noalign{\hrule height 1pt}
        \end{tabular}
    }
    \caption{\sf Cross-validation performance metrics of the SweetDeep model. $\left[\uparrow\right]$ indicates that `higher is better,' while $\left[\downarrow\right]$ means `lower is better.' Best values are shown in bold. On an average, a patient has 20 instances, enabling more informed patient-level predictions of type 2 diabetic status compared to instance-level classification alone.}
    \label{tab:sweetdeep-performance}
\end{table*}

\begin{figure*}[ht]
    \centering
    \begin{subfigure}[b]{0.48\textwidth}
        \centering
        \begin{tikzpicture}[scale=0.85]
            \begin{axis}[
                xlabel=\textit{Mean predicted $\hat{p}_x$},
                ylabel=\textit{Fraction of positives},
                grid=both,
                grid style={line width=0.5mm, draw=gray!3},
                xmin=0, xmax=1.0,
                ymin=0, ymax=1.0,
                xtick={0,0.2,...,1.0},
                ytick={0,0.2,...,1.0},
                axis equal image
            ]
            \addplot[mark=*,Periwinkle,line width=0.8mm,mark size=1,opacity=1] plot coordinates {
                (0.046271728417696036, 0.07767624020887728)
                (0.1439367656897903, 0.08115183246073299)
                (0.24661729946695102, 0.18336886993603413)
                (0.34968431207602346, 0.260233918128655)
                (0.44972777930091223, 0.3829787234042553)
                (0.5511110473666659, 0.51)
                (0.6509162083050847, 0.576271186440678)
                (0.7522446436641221, 0.7048346055979644)
                (0.8566974605519488, 0.823051948051948)
                (0.9387662253201956, 0.8555008210180624)
            };
            \addplot[dotted, gray, line width=0.4mm] plot coordinates {
                (0, 0)
                (1.0, 1.0)
            };
            \end{axis}
        \end{tikzpicture}
        \caption{\sf \textcolor{Periwinkle}{Instance-level} calibration (ECE = 5.4\%)}
    \end{subfigure}
    \hfill
    \begin{subfigure}[b]{0.48\textwidth}
        \centering
        \begin{tikzpicture}[scale=0.85]
            \begin{axis}[
                xlabel=\textit{Mean predicted $\hat{p}_X$},
                ylabel=\textit{Fraction of positives},
                grid=both,
                grid style={line width=0.5mm, draw=gray!3},
                xmin=0, xmax=1.0,
                ymin=0, ymax=1.0,
                xtick={0,0.2,...,1.0},
                ytick={0,0.2,...,1.0},
                axis equal image
            ]
            \addplot[mark=*,Fuchsia,line width=0.8mm,mark size=1,opacity=1] plot coordinates {
                (0.0488366413068318, 0.07352941176470588)
                (0.13363696816108622, 0.08333333333333333)
                (0.24859515657900733, 0.20689655172413793)
                (0.3499922810312358, 0.4666666666666667)
                (0.46187083150391833, 0.2857142857142857)
                (0.5586694164401831, 0.65)
                (0.6417722780835273, 0.7777777777777778)
                (0.7651164610615385, 0.6875)
                (0.8561863086462577, 0.8536585365853658)
                (0.9315547811114037, 0.8928571428571429)
            };
            \addplot[dotted, gray, line width=0.4mm] plot coordinates {
                (0, 0)
                (1.0, 1.0)
            };
            \end{axis}
        \end{tikzpicture}
        \caption{\sf \textcolor{Fuchsia}{Patient-level} calibration (ECE = 5.5\%)}
    \end{subfigure}

    \caption{\sf SweetDeep three-fold calibration curves. Points closer to the dotted diagonal line indicate better bin-wise calibration.}
    \label{fig:calibration}
\end{figure*}
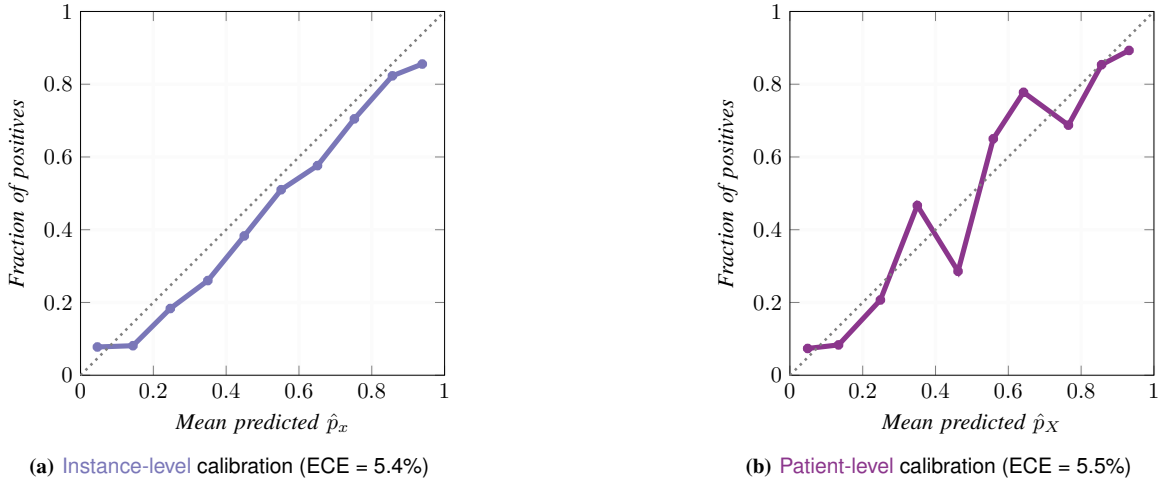

\subsection{Model Performance} \label{perf}

Table~\ref{tab:sweetdeep-performance} provides a summary of SweetDeep cross-validation performance, at both the instance and patient levels. At the instance level, SweetDeep achieves an accuracy of 80.9\% and an $\mathcal{F}_1$-score of 79.8\%, indicating balanced classification performance. The model exhibits higher specificity (82.9\%) than sensitivity (77.5\%), reflecting a conservative bias toward ND predictions. This behavior reduces the likelihood of `false alarm' outcomes but slightly lowers sensitivity to true T2D cases. Adjusting the prediction threshold from 50\% enables a trade-off between sensitivity and specificity (see Sec. \ref{threshold}), although accuracy and $\mathcal{F}_1$ scores are affected accordingly.

Averaging predictions across all instances per subject (i.e., labeling based on $\hat{p}_X \ge 0.5$) results in higher patient-level performance, with accuracy increasing to 82.5\% and specificity to 84.6\%. These results indicate that integrating multiple recordings per individual -- across various days and times -- enhances performance compared to isolated instance-level classification. At the patient level, $\mathcal{F}_1$ (82.1\%) closely matches accuracy, further demonstrating that SweetDeep maintains class-balanced classification performance.

Calibration results are presented in Fig.~\ref{fig:calibration}. Both instance- and patient-level predictions exhibit well-calibrated probability estimates, with expected calibration errors of 5.4\% and 5.5\%, respectively. In both cases, the empirical fraction of positives tracks the dotted diagonal line, indicating that predicted probabilities align well with observed bin-wise frequencies. 

Overall, these findings support the conclusion that SweetDeep provides reasonably accurate and well-calibrated predictions of T2D likelihood, by enabling both rapid instance-level screening and robust patient-level assessment, depending on the desired timescale of smartwatch-based inference.

\begin{table*}[ht]
    \normalsize
    \centering
    \makebox[\textwidth][c]{
        \begin{tabular}{!{\vrule width 1pt}c!{\vrule width 1pt}c!{\vrule width 1pt}c!{\vrule width 1pt}c!{\vrule width 1pt}c!{\vrule width 1pt}c!{\vrule width 1pt}c!{\vrule width 1pt}}
        \noalign{\hrule height 1pt}
        \textbf{Category} & \textbf{Ablation / Variation} & \textbf{Acc~(\%)} $\left[\uparrow\right]$ & \textbf{Sens~(\%)} $\left[\uparrow\right]$ & \textbf{Spec~(\%)} $\left[\uparrow\right]$ & $\boldsymbol{\mathcal{F}_1}$~\textbf{(\%)} $\left[\uparrow\right]$ &
        \textbf{ECE~(\%)} $\left[\downarrow\right]$ \Tstrut\Bstrut \\
        \noalign{\hrule height 1pt}
        \multirow{1}{*}{Architectural} & No Dropout & \textcolor{gray}{82.1} & \textcolor{gray}{78.9} & 84.6 & \textcolor{gray}{81.7} & 5.2 \Tstrut\Bstrut \\
        \cline{2-7}
        & No BatchNorm & \textcolor{gray}{81.4} & \textcolor{gray}{78.9} & \textcolor{gray}{83.3} & \textcolor{gray}{81.1} & 5.1 \Tstrut\Bstrut \\
        \cline{2-7}
        & No DO + No BN & \textcolor{gray}{81.1} & \textcolor{gray}{78.0} & \textcolor{gray}{83.3} & \textcolor{gray}{80.7} & \textbf{\textcolor{Fuchsia}{2.8}} \Tstrut\Bstrut \\
        \noalign{\hrule height 1pt}
        \multirow{1}{*}{Model Size} & Wider Layers & \textcolor{gray}{82.1} & \textcolor{gray}{76.4} & 86.4 & \textcolor{gray}{81.6} & 5.2  \Tstrut\Bstrut \\
        \cline{2-7}
        & Deeper Network & \textcolor{gray}{81.4} & \textcolor{gray}{78.0} & \textcolor{gray}{84.0} & \textcolor{gray}{81.0} & \textcolor{gray}{6.4}  \Tstrut\Bstrut \\
        \cline{2-7}
        & Wider + Deeper & \textcolor{gray}{81.1} & \textcolor{gray}{74.8} & 85.8 & \textcolor{gray}{80.5} & \textcolor{gray}{6.0} \Tstrut\Bstrut \\
        \noalign{\hrule height 1pt}
        \multirow{1}{*}{Feature} & No PPG-BP & \textcolor{gray}{81.1} & \textcolor{gray}{78.9} & \textcolor{gray}{82.7} & \textcolor{gray}{80.7} & 3.8 \Tstrut\Bstrut \\
        \cline{2-7}
        & No Family History & \textcolor{gray}{81.1} & 79.7 & \textcolor{gray}{82.1} & \textcolor{gray}{80.8} & 4.7  \Tstrut\Bstrut \\
        \cline{2-7}
        & No ECG & \textcolor{gray}{80.4} & \textcolor{gray}{78.9} & \textcolor{gray}{81.5} & \textcolor{gray}{80.0} & 5.5  \Tstrut\Bstrut \\
        \cline{2-7}
        & No Time & \textcolor{gray}{79.3} & \textcolor{gray}{77.2} & \textcolor{gray}{80.9} & \textcolor{gray}{79.0} & 4.0  \Tstrut\Bstrut \\
        \cline{2-7}
        & No BIA & \textcolor{gray}{79.3} & \textcolor{gray}{78.9} & \textcolor{gray}{79.6} & \textcolor{gray}{79.0} & \textcolor{gray}{7.3}  \Tstrut\Bstrut \\
        \cline{2-7}
        & No Age & \textcolor{gray}{75.4} & \textcolor{gray}{70.7} & \textcolor{gray}{79.0} & \textcolor{gray}{74.9} & \textcolor{gray}{8.6}  \Tstrut\Bstrut \\
        \noalign{\hrule height 1pt}
        \multirow{1}{*}{Threshold} & 40\% & \textcolor{gray}{80.4} & \textbf{\textcolor{Fuchsia}{82.9}} & \textcolor{gray}{78.4} & \textcolor{gray}{80.2} & 5.5 \Tstrut\Bstrut \\
        \cline{2-7}
        & 60\% & \textcolor{gray}{80.4} & \textcolor{gray}{69.1} & \textbf{\textcolor{Fuchsia}{88.9}} & \textcolor{gray}{79.5} & 5.5 \Tstrut\Bstrut \\
        \noalign{\hrule height 1pt}
        \multirow{1}{*}{None} & SweetDeep & \textbf{\textcolor{Fuchsia}{82.5}} & 79.7 & 84.6 & \textbf{\textcolor{Fuchsia}{82.1}} & 5.5 \Tstrut\Bstrut \\
        \noalign{\hrule height 1pt}
        \end{tabular}
    }
    \caption{\sf Patient-level results after various ablations / variations of SweetDeep. $\left[\uparrow\right]$ indicates that `higher is better,' while $\left[\downarrow\right]$ means `lower is better.' Best metric values are in \textbf{\textcolor{Fuchsia}{bolded purple}}, while worse values compared to the original SweetDeep are in \textcolor{gray}{light gray}.}
    \label{tab:sweetdeep-ablations}
\end{table*}

\subsection{Architectural Ablations}

Table~\ref{tab:sweetdeep-ablations} shows the effects of various ablations and training variations on the patient-level performance of SweetDeep.

The \textbf{architectural ablations} selectively remove DO and BN layers to assess their contribution to performance. Both DO and BN serve as regularizers, mitigating overfitting and stabilizing training. In the ``No DO + No BN'' ablation, all accuracy-related metrics degrade, indicating that these layers improve test performance. Similar but smaller declines occur when either DO or BN is removed individually, suggesting that both contribute to model generalization.

Conversely, calibration improves when one or both of these layers are removed, as reflected by lower ECE. This trend aligns with prior findings that BN~\cite{guo2017calibration} and DO~\cite{xie2022inference} can increase miscalibration due to train-test discrepancies in layer activation patterns. The notably low ECE (2.8\%) observed for the ``No DO + No BN'' model suggests more reliable probability estimates, potentially allowing for trustworthy, continuous communication of diabetic risk to users and clinicians after aggregating multiple sensor recordings.

\subsection{Model Size Variations} \label{size-var}

Next, \textbf{model size variations} evaluate how increasing network width and depth affects test performance. Larger models are often associated with degraded calibration \cite{guo2017calibration}; hence, this experiment tests this hypothesis with SweetDeep.

The baseline configuration has layer widths [35, 64, 8, 2]. Three variants are examined:
\vspace{0.3em}
\begin{itemize}
    \item \textbf{Wider Layers}: [35, 128, 16, 2]
    \begin{itemize}[label=\customarrowbullet]
        \item Each hidden layer doubles in size.
    \end{itemize}
    \item \textbf{Deeper Network}: [35, 64, 64, 8, 2]
    \begin{itemize}[label=\customarrowbullet]
        \item The first hidden layer is duplicated.
    \end{itemize}
    \item \textbf{Wider + Deeper}: [35, 128, 128, 16, 2]
    \begin{itemize}[label=\customarrowbullet]
        \item Combines both ``Wider Layers'' and ``Deeper Network.''
    \end{itemize}
\end{itemize}
\vspace{0.3em}

As shown in Table~\ref{tab:sweetdeep-ablations}, the ``Wider + Deeper'' and ``Deeper Network'' models exhibit slightly higher ECE values, consistent with expectations that larger architectures reduce calibration quality. The ``Wider Layers'' variant, however, shows a marginal improvement in ECE (by 0.3\%). 

These calibration effects (within $\pm$1\% ECE) are modest and unlikely to meaningfully affect probabilistic reliability. Nonetheless, all larger models show lower accuracy and $\mathcal F_1$, suggesting increased overfitting without compensatory gains in calibration. Hence, expanding SweetDeep’s size beyond the baseline configuration appears suboptimal for this dataset.

\subsection{Feature Ablations}

\textbf{Feature ablations} selectively eliminate each related group of features (e.g., ECG) to infer the contribution of each feature set to classification accuracy and calibration. As can be seen from Table~\ref{tab:sweetdeep-ablations}, every feature ablation results in accuracy and $\mathcal F_1$ losses -- some more drastic than others -- with similar overall decreases in sensitivity and specificity. This result indicates that every feature group contributes meaningfully to smartwatch-based classification of T2D.

Effects of these ablations on model calibration are less clear-cut, with some feature groups (PPG-BP and Time) decreasing ECE by $>$1\% when removed, suggesting that these features appear detrimental to calibration. This may be due to factors such as sensor or timing fluctuations, or the use of SMOTE (a class rebalancing technique during training) disrupting the feature distribution between training and testing.

The \textbf{Age} feature leads to the largest accuracy decline when dropped, suggesting that it is the most ``important'' feature. Indeed, the risk of elevated glucose and type 2 diabetes in the general population increases progressively with age \cite{suastika2012age, fletcher2002risk}. The next most influential feature group, BIA, estimates a patient's metabolic profile (including body fat and metabolic rate), which is closely linked to diabetes risk \cite{singh2025type}. Time encodings are also important, as they allow sensor information to be contextualized by time of day, capturing the influence of circadian rhythms on physiological readings.

\begin{figure*}[!htbp]
    \centering
    \includegraphics[width=\textwidth]{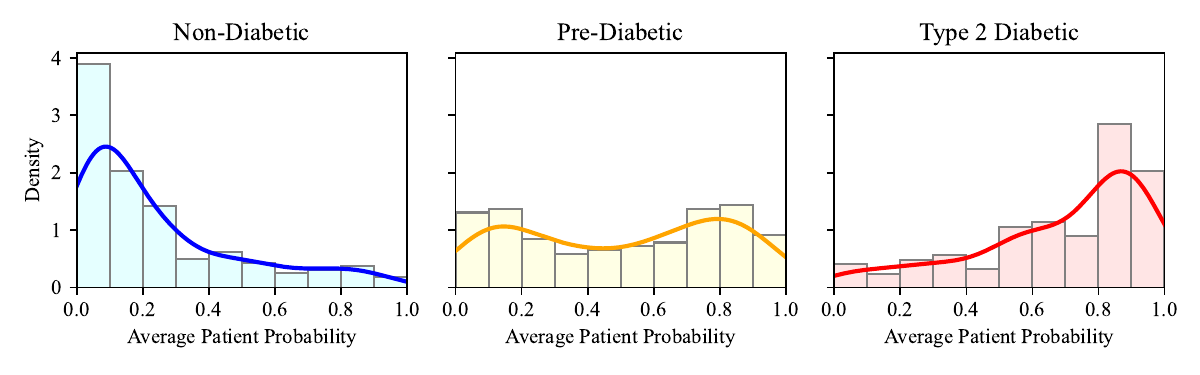}
\caption{\sf Distributions of test-time patient-level predictions by SweetDeep for ND, PD, and T2D cohorts. PD probabilities are obtained from a model trained on the full ND vs.~T2D dataset. While ND and T2D distributions are concentrated in regions of high confidence, the PD distribution is bimodal and flatter, reflecting greater ambiguity.}

    \label{fig:dists}
\end{figure*}

\begin{table*}[ht]
    \normalsize
    \centering
    \makebox[\textwidth][c]{%
        \begin{tabular}{!{\vrule width 1pt}c!{\vrule width 1pt}c!{\vrule width 1pt}c!{\vrule width 1pt}c!{\vrule width 1pt}c!{\vrule width 1pt}c!{\vrule width 1pt}}
        \noalign{\hrule height 1pt}
        \textbf{Coverage~(\%)} & \textbf{Accuracy~(\%)} $\left[\uparrow\right]$ & \textbf{Sensitivity~(\%)} $\left[\uparrow\right]$ &
        \textbf{Specificity~(\%)} $\left[\uparrow\right]$ & $\boldsymbol{\mathcal{F}_1}$~$\textbf{(\%)}$ $\left[\uparrow\right]$ &
        \textbf{ECE~(\%)} $\left[\downarrow\right]$ \Tstrut\Bstrut \\
        \noalign{\hrule height 1pt}
        90.5 & 84.5 & 81.1 & 87.1 & 84.2 & 5.5 \Tstrut\Bstrut \\
        \noalign{\hrule height 1pt}
        \end{tabular}
    }
    \caption{\sf Patient-level performance metrics, when \textbf{abstention} is applied for probabilities between 42\% and 58\%. Note that due to decreased dataset coverage, these results cannot be directly compared with the baseline SweetDeep model's performance.}
    \label{tab:sweetdeep-abstention}
\end{table*}

\subsection{Thresholding Variations} \label{threshold}

Table~\ref{tab:sweetdeep-ablations} also presents two \textbf{thresholding variations}, which examine the impact of adjusting the probability threshold on performance. SweetDeep uses a standard threshold of 50\%, reflecting equal weighting of ND and T2D predictions. Here, the same model is evaluated at alternative thresholds, without retraining or recalibrating (hence ECE stays the same).

Changing the threshold trades sensitivity for specificity. At a 60\% threshold, the model is less likely to predict T2D, resulting in higher specificity (88.9\%) but lower sensitivity (69.1\%). Conversely, lowering the threshold to 40\% increases sensitivity to 82.9\% while decreasing specificity to 78.4\%.

Both threshold adjustments lead to more than 2\% reductions in accuracy and $\mathcal F_1$ score. Since SweetDeep is well-calibrated, manipulating the threshold would flip more correct predictions than it fixes wrong ones. Therefore, unless targeting a specific objective (e.g., minimizing false alarms), adjusting the threshold from the default 50\% provides minimal improvement and may reduce diabetes screening performance.

\subsection{Cohort-Specific Trends}

SweetDeep is trained and tested on ND and T2D data, but it is also informative to examine predictions for individuals with PD. PD patients may exhibit early physiological changes associated with elevated glucose, including subtle alterations in cardiac function detectable by ECG, but they are not yet fully diabetic. Consequently, they may be assigned to either class by models like SweetDeep.

To assess model behavior on PD patients, who are not included in the cross-validation folds, we train a version of SweetDeep on all ND and T2D patients using the same hyperparameters. The resulting prediction distribution is graphed in Fig.~\ref{fig:dists}, with histograms and kernel density plots from ND and T2D cross-validation also included as a reference.

As expected, most ND patients have predicted probabilities below 50\%, while most T2D patients have probabilities above 50\%, consistent with prior thresholding analysis (Section~\ref{threshold}). Both ND and T2D distributions are unimodal, with modes near high-confidence regions ($\hat p \approx 0$ or $\hat p \approx 1$, respectively). In contrast, the PD distribution is bimodal and flatter, with peaks near both extremes, indicating that PD patients are not consistently classified as either ND or T2D.

Future work may investigate a three-class model trained on data from ND, PD, and T2D patients. Such an extension may require more informative or additional sensors, as PD is less readily characterized using current biophysical signals. 

\subsection{Abstention Results}

Finally, we propose an abstention adapter applied after the softmax layer, requiring thresholding with no additional model training. This modification trades reduced dataset coverage for improved credibility by allowing the model to explicitly defer uncertain predictions. Leveraging SweetDeep’s well-calibrated probabilities, patients with predictions between 42\% and 58\% are predicted as ``Don't Know.'' This range, centered on the 50\% threshold, was empirically selected to abstain on \textit{at most 10\%} of patients during cross-validation.

As shown in Table~\ref{tab:sweetdeep-abstention}, accuracy-related metrics among retained predictions appear higher, reflecting the exclusion of 27 low-confidence patients from the dataset during evaluation. However, this does not constitute a genuine gain in model capability, as abstention reduces coverage and consequently leads to fewer correct detections of T2D or ND. Nevertheless, abstention can enhance the model’s perceived trustworthiness by explicitly acknowledging predictive uncertainty.

If abstention is employed in practice, users or clinicians should be clearly informed that follow-up biochemical tests (e.g., FPG or OGTT) are necessary to accurately determine the diabetic status of uncommon ``Don't Know" cases.

\section{Conclusion} \label{conclusion}

In this article, we introduced SweetDeep, a compact neural network with fewer than 3,000 parameters, trained to detect T2D using smartwatch sensing and questionnaire data alone. Data collection was conducted through a decentralized clinical trial in participants’ own homes, closely mirroring real-world deployment conditions. By integrating cardiovascular, metabolic, demographic, genetic, and temporal features, SweetDeep achieves accurate (82.5\%) and well-calibrated predictions, demonstrating the feasibility of wearable-based diabetes screening as a precursor to standard clinical testing. With continued validation across larger and more diverse populations, SweetDeep could enable accessible and reliable diabetes detection in everyday settings, helping to lower barriers to timely diagnosis and support proactive care.

\begingroup
  \small
  \bibliographystyle{IEEEtran}
  \bibliography{refs}

@article{singh2025type,
  title={Type 2 diabetes mellitus: A comprehensive review of pathophysiology, comorbidities, and emerging therapies},
  author={Singh, Aditi and Shadangi, Sucharita and Gupta, Pulkit Kr and Rana, Soumendra},
  journal={Comprehensive Physiology},
  volume={15},
  number={1},
  pages={e70003},
  year={2025},
  publisher={Wiley Online Library}
}

@online{cdc_diabetes_statistics,
  author       = {{Centers for Disease Control and Prevention}},
  title        = {A Report Card: Diabetes in the United States Infographic},
  year         = {2024},
  month        = may,
  url          = {https://www.cdc.gov/diabetes/communication-resources/diabetes-statistics.html},
  note         = {Accessed: 2025‑09‑27}
}

@article{yin2019diabdeep,
  title={DiabDeep: Pervasive diabetes diagnosis based on wearable medical sensors and efficient neural networks},
  author={Yin, Hongxu and Mukadam, Bilal and Dai, Xiaoliang and Jha, Niraj K.},
  journal={IEEE Transactions on Emerging Topics in Computing},
  volume={9},
  number={3},
  pages={1139--1150},
  year={2019},
  publisher={IEEE}
}

@article{tucker2020limited,
  title={Limited agreement between classifications of diabetes and prediabetes resulting from the {OGTT}, hemoglobin {A1c}, and fasting glucose tests in 7412 {US} adults},
  author={Tucker, Larry A},
  journal={Journal of Clinical Medicine},
  volume={9},
  number={7},
  pages={2207},
  year={2020},
  publisher={MDPI}
}

@article{american20242,
  title={2. Diagnosis and classification of diabetes: Standards of care in diabetes—2024},
  author={American Diabetes Association Professional Practice Committee},
  journal={Diabetes Care},
  volume={47},
  number={Supplement\_1},
  pages={S20--S42},
  year={2024},
  publisher={American Diabetes Association}
}

@misc{idf2025,
  author       = {{International Diabetes Federation}},
  title        = {{IDF Global Clinical Practice Recommendations for Managing Type 2 Diabetes – 2025}},
  year         = {2025},
  howpublished = {\url{https://idf.org/t2d-cpr-2025}},
  note         = {Accessed: 2025-09-28}
}

@article{teuscher1989diabetes,
  title={Diabetes and hypertension: Blood pressure in clinical diabetic patients and a control population},
  author={Teuscher, Arthur and Egger, Matthias and Herman, Joseph B.},
  journal={Archives of Internal Medicine},
  volume={149},
  number={9},
  pages={1942--1945},
  year={1989},
  publisher={American Medical Association}
}

@article{benichou2018heart,
  title={Heart rate variability in type 2 diabetes mellitus: A systematic review and meta--analysis},
  author={Benichou, Thomas and Pereira, Bruno and Mermillod, Martial and Tauveron, Igor and Pfabigan, Daniela and Maqdasy, Salwan and Dutheil, Frederic},
  journal={PloS One},
  volume={13},
  number={4},
  pages={e0195166},
  year={2018},
  publisher={Public Library of Science San Francisco, CA USA}
}

@article{cordeiro2021hyperglycemia,
  title={Hyperglycemia identification using {ECG} in deep learning era},
  author={Cordeiro, Renato and Karimian, Nima and Park, Younghee},
  journal={Sensors},
  volume={21},
  number={18},
  pages={6263},
  year={2021},
  publisher={MDPI}
}

@article{susana2022non,
  title={Non-invasive classification of blood glucose level for early detection diabetes based on photoplethysmography signal},
  author={Susana, Ernia and Ramli, Kalamullah and Murfi, Hendri and Apriantoro, Nursama Heru},
  journal={Information},
  volume={13},
  number={2},
  pages={59},
  year={2022},
  publisher={MDPI}
}

@article{zanelli2022diabetes,
  title={Diabetes detection and management through photoplethysmographic and electrocardiographic signals analysis: A systematic review},
  author={Zanelli, Serena and Ammi, Mehdi and Hallab, Magid and El Yacoubi, Mounim A.},
  journal={Sensors},
  volume={22},
  number={13},
  pages={4890},
  year={2022},
  publisher={MDPI}
}

@article{piet2023non,
  title={Non-invasive wearable devices for monitoring vital signs in patients with type 2 diabetes mellitus: A systematic review},
  author={Piet, Artur and Jablonski, Lennart and Daniel Onwuchekwa, Jennifer I. and Unkel, Steffen and Weber, Christian and Grzegorzek, Marcin and Ehlers, Jan P. and Gaus, Olaf and Neumann, Thomas},
  journal={Bioengineering},
  volume={10},
  number={11},
  pages={1321},
  year={2023},
  publisher={MDPI}
}

@article{moses2023non,
  title={Non-invasive blood glucose monitoring technology in diabetes management},
  author={Moses, Jeban Chandir and Adibi, Sasan and Wickramasinghe, Nilmini and Nguyen, Lemai and Angelova, Maia and Islam, Sheikh Mohammed Shariful},
  journal={Mhealth},
  volume={10},
  pages={9},
  year={2023}
}

@incollection{vashist2019commercially,
  title={Commercially available smartphone-based personalized mobile healthcare technologies},
  author={Vashist, Sandeep Kumar and Luong, John H. T.},
  booktitle={Point-of-Care Technologies Enabling Next-Generation Healthcare Monitoring and Management},
  pages={81--115},
  year={2019},
  publisher={Springer}
}

@article{bergman2024international,
  title={International Diabetes Federation Position Statement on the 1-hour post-load plasma glucose for the diagnosis of intermediate hyperglycaemia and type 2 diabetes},
  author={Bergman, Michael and Manco, Melania and Satman, Ilhan and Chan, Juliana and Schmidt, Maria In{\^e}s and Sesti, Giorgio and Fiorentino, Teresa Vanessa and Abdul-Ghani, Muhammad and Jagannathan, Ram and Aravindakshan, Pramod Kumar Thyparambil and others},
  journal={Diabetes Research and Clinical Practice},
  volume={209},
  pages={111589},
  year={2024},
  publisher={Elsevier}
}

@article{nathan2007relationship,
  title={Relationship between glycated haemoglobin levels and mean glucose levels over time},
  author={Nathan, D M. and Turgeon, H. and Regan, S.},
  journal={Diabetologia},
  volume={50},
  number={11},
  pages={2239--2244},
  year={2007},
  publisher={Springer}
}

@article{pop2010cardiac,
  title={Cardiac autonomic neuropathy in diabetes: A clinical perspective},
  author={Pop-Busui, Rodica},
  journal={Diabetes Care},
  volume={33},
  number={2},
  pages={434},
  year={2010}
}

@article{duque2021cardiovascular,
  title={Cardiovascular autonomic neuropathy in diabetes: Pathophysiology, clinical assessment and implications},
  author={Duque, Alice and Mediano, Mauro Felippe Felix and De Lorenzo, Andrea and Rodrigues Jr., Luiz Fernando},
  journal={World Journal of Diabetes},
  volume={12},
  number={6},
  pages={855},
  year={2021}
}

@article{azad1999effects,
  title={The effects of intensive glycemic control on neuropathy in the {VA} cooperative study on type {II} diabetes mellitus {(VA CSDM)}},
  author={Azad, Nasrin and Emanuele, Nicholas V. and Abraira, Carlos and Henderson, William G. and Colwell, John and Levin, Seymore R. and Nuttall, Frank Q. and Comstock, John P. and Sawin, Clark T. and Silbert, Cynthia and others},
  journal={Journal of Diabetes and its Complications},
  volume={13},
  number={5-6},
  pages={307--313},
  year={1999},
  publisher={Elsevier}
}

@article{chaibub2019detecting,
  title={Detecting the impact of subject characteristics on machine learning-based diagnostic applications},
  author={Chaibub Neto, Elias and Pratap, Abhishek and Perumal, Thanneer M. and Tummalacherla, Meghasyam and Snyder, Phil and Bot, Brian M. and Trister, Andrew D. and Friend, Stephen H. and Mangravite, Lara and Omberg, Larsson},
  journal={NPJ Digital Medicine},
  volume={2},
  number={1},
  pages={99},
  year={2019},
  publisher={Nature Publishing Group UK London}
}

@article{tougui2021impact,
  title={Impact of the choice of cross-validation techniques on the results of machine learning-based diagnostic applications},
  author={Tougui, Ilias and Jilbab, Abdelilah and El Mhamdi, Jamal},
  journal={Healthcare Informatics Research},
  volume={27},
  number={3},
  pages={189--199},
  year={2021},
  publisher={Korean Society of Medical Informatics}
}

@article{castaneda2018review,
  title={A review on wearable photoplethysmography sensors and their potential future applications in health care},
  author={Castaneda, Denisse and Esparza, Aibhlin and Ghamari, Mohammad and Soltanpur, Cinna and Nazeran, Homer},
  journal={International Journal of Biosensors \& Bioelectronics},
  volume={4},
  number={4},
  pages={195},
  year={2018}
}

@article{brands1996poor,
  title={Poor glycemic control induces hypertension in diabetes mellitus},
  author={Brands, Michael W. and Hopkins, Timothy E.},
  journal={Hypertension},
  volume={27},
  number={3},
  pages={735--739},
  year={1996},
  publisher={Lippincott Williams \& Wilkins}
}

@article{wu2022association,
  title={Association of mean arterial pressure with 5-year risk of incident diabetes in Chinese adults: A secondary population-based cohort study},
  author={Wu, Yang and Hu, Haofei and Cai, Jinlin and Chen, Runtian and Zuo, Xin and Cheng, Heng and Yan, Dewen},
  journal={BMJ Open},
  volume={12},
  number={9},
  pages={e048194},
  year={2022},
  publisher={British Medical Journal Publishing Group}
}

@inproceedings{guo2017calibration,
  title={On calibration of modern neural networks},
  author={Guo, Chuan and Pleiss, Geoff and Sun, Yu and Weinberger, Kilian Q.},
  booktitle={International Conference on Machine Learning},
  pages={1321--1330},
  year={2017},
  organization={PMLR}
}

@article{ovadia2019can,
  title={Can you trust your model's uncertainty? Evaluating predictive uncertainty under dataset shift},
  author={Ovadia, Yaniv and Fertig, Emily and Ren, Jie and Nado, Zachary and Sculley, David and Nowozin, Sebastian and Dillon, Joshua and Lakshminarayanan, Balaji and Snoek, Jasper},
  journal={Advances in Neural Information Processing systems},
  volume={32},
  year={2019}
}

@article{elming2002qtc,
  title={{QTc} interval in the assessment of cardiac risk},
  author={Elming, Hanne and Brendorp, Bente and K{\o}ber, Lars and Sahebzadah, Najia and Torp-Petersen, Christian},
  journal={Cardiac Electrophysiology Review},
  volume={6},
  number={3},
  pages={289--294},
  year={2002},
  publisher={Springer}
}

@article{electrophysiology1996heart,
  title={Heart rate variability: Standards of measurement, physiological interpretation, and clinical use},
  author={Electrophysiology, Task Force of the European Society of Cardiology the North American Society of Pacing},
  journal={Circulation},
  volume={93},
  number={5},
  pages={1043--1065},
  year={1996},
  publisher={Lippincott Williams \& Wilkins}
}

@article{vandenberk2016qt,
  title={Which {QT} correction formulae to use for {QT} monitoring?},
  author={Vandenberk, Bert and Vandael, Eline and Robyns, Tomas and Vandenberghe, Joris and Garweg, Christophe and Foulon, Veerle and Ector, Joris and Willems, Rik},
  journal={Journal of the American Heart Association},
  volume={5},
  number={6},
  pages={e003264},
  year={2016}
}

@article{baumgart1991circadian,
  title={Circadian rhythm of blood pressure: Internal and external time triggers},
  author={Baumgart, Peter},
  journal={Chronobiology International},
  volume={8},
  number={6},
  pages={444--450},
  year={1991},
  publisher={Taylor \& Francis}
}

@article{huikuri1990reproducibility,
  title={Reproducibility and circadian rhythm of heart rate variability in healthy subjects},
  author={Huikuri, Heikki V. and Kessler, Kenneth M. and Terracall, Elisabeth and Castellanos, Agustin and Linnaluoto, Markku K. and Myerburg, Robert J.},
  journal={The American Journal of Cardiology},
  volume={65},
  number={5},
  pages={391--393},
  year={1990},
  publisher={Elsevier}
}

@article{vaswani2017attention,
  title={Attention is all you need},
  author={Vaswani, Ashish and Shazeer, Noam and Parmar, Niki and Uszkoreit, Jakob and Jones, Llion and Gomez, Aidan N. and Kaiser, {\L}ukasz and Polosukhin, Illia},
  journal={Advances in Neural Information Processing Systems},
  volume={30},
  year={2017}
}

@article{bansal2025temporal,
  title={Temporal encoding strategies for energy time series prediction},
  author={Bansal, Aayam and Balaji, Keertan and Lalani, Zeus},
  journal={arXiv preprint arXiv:2503.15456},
  year={2025}
}

@article{eddin2023location,
  title={Location-aware adaptive normalization: A deep learning approach for wildfire danger forecasting},
  author={Eddin, Mohamad Hakam Shams and Roscher, Ribana and Gall, Juergen},
  journal={IEEE Transactions on Geoscience and Remote Sensing},
  volume={61},
  pages={1--18},
  year={2023},
  publisher={IEEE}
}

@article{srivastava2014dropout,
  title={Dropout: A simple way to prevent neural networks from overfitting},
  author={Srivastava, Nitish and Hinton, Geoffrey and Krizhevsky, Alex and Sutskever, Ilya and Salakhutdinov, Ruslan},
  journal={The Journal of Machine Learning Research},
  volume={15},
  number={1},
  pages={1929--1958},
  year={2014},
  publisher={JMLR. org}
}

@inproceedings{ioffe2015batch,
  title={Batch normalization: Accelerating deep network training by reducing internal covariate shift},
  author={Ioffe, Sergey and Szegedy, Christian},
  booktitle={International Conference on Machine Learning},
  pages={448--456},
  year={2015}
}

@article{kingma2014adam,
  title={Adam: A method for stochastic optimization},
  author={Kingma, Diederik P.},
  journal={arXiv preprint arXiv:1412.6980},
  year={2014}
}

@article{chawla2002smote,
  title={{SMOTE}: Synthetic minority over-sampling technique},
  author={Chawla, Nitesh V. and Bowyer, Kevin W. and Hall, Lawrence O. and Kegelmeyer, W. Philip},
  journal={Journal of Artificial Intelligence Research},
  volume={16},
  pages={321--357},
  year={2002}
}

@article{van2021decentralized,
  title={Decentralized clinical trials: The future of medical product development?},
  author={Van Norman, Gail A.},
  journal={Basic to Translational Science},
  volume={6},
  number={4},
  pages={384--387},
  year={2021},
  publisher={American College of Cardiology Foundation Washington DC}
}

@article{herbei2006classification,
  title={Classification with reject option},
  author={Herbei, Radu and Wegkamp, Marten H.},
  journal={The Canadian Journal of Statistics/La Revue Canadienne de Statistique},
  pages={709--721},
  year={2006},
  publisher={JSTOR}
}

@article{opitz2019macro,
  title={Macro {F1} and macro {F1}},
  author={Opitz, Juri and Burst, Sebastian},
  journal={arXiv preprint arXiv:1911.03347},
  year={2019}
}

@inproceedings{xie2022inference,
  title={Inference calibration of neural network models},
  author={Xie, Xiaohu and Madankan, Reza and Wu, Kairui},
  booktitle={IEEE International Conference on Knowledge Graph},
  pages={347--354},
  year={2022},
  organization={IEEE}
}

@incollection{suastika2012age,
  title={Age is an important risk factor for type 2 diabetes mellitus and cardiovascular diseases},
  author={Suastika, Ketut and Dwipayana, Pande and Semadi, Made Siswadi and Kuswardhani, R. A. Tuty},
  booktitle={Glucose Tolerance},
  year={2012},
  publisher={IntechOpen}
}

@article{fletcher2002risk,
  title={Risk factors for type 2 diabetes mellitus},
  author={Fletcher, Barbara and Gulanick, Meg and Lamendola, Cindy},
  journal={Journal of Cardiovascular Nursing},
  volume={16},
  number={2},
  pages={17--23},
  year={2002},
  publisher={LWW}
}
\endgroup

\end{document}